\documentclass[10pt,twocolumn, letterpaper]{article}

\usepackage[pagenumbers]{cvpr} 

\usepackage{graphicx}
\usepackage{amsmath}
\usepackage{amssymb}
\usepackage{booktabs}

\usepackage[pagebackref,breaklinks,colorlinks]{hyperref}

\usepackage[capitalize]{cleveref}
\crefname{section}{Sec.}{Secs.}
\Crefname{section}{Section}{Sections}
\Crefname{table}{Table}{Tables}
\crefname{table}{Tab.}{Tabs.}

\begin{document}

\title{PP-MobileSeg: Explore the Fast and Accurate Semantic Segmentation Model on Mobile Devices}

\author{Shiyu Tang$^*$ \qquad Ting Sun$^*$ \qquad Juncai Peng$^*$ \qquad Guowei Chen \qquad Yuying Hao \\ \qquad Manhui Lin 
\qquad Zhihong Xiao \qquad Jiangbin You \qquad Yi Liu  \\Baidu Inc.\\
{\tt\small \{tangshiyu, sunting13, pengjuncai\}@baidu.com}}

\maketitle
\def\thefootnote{*}\footnotetext{These authors contributed equally to this work}\def\thefootnote{\arabic{footnote}}

\begin{abstract}
The success of transformers in computer vision has led to several attempts to adapt them for mobile devices, but their performance remains unsatisfactory in some real-world applications. To address this issue, we propose PP-MobileSeg, a semantic segmentation model that achieves state-of-the-art performance on mobile devices. PP-MobileSeg comprises three novel parts: the StrideFormer backbone, the Aggregated Attention Module (AAM), and the Valid Interpolate Module (VIM). The four-stage StrideFormer backbone is built with MV3 blocks and strided SEA attention, and it is able to extract rich semantic and detailed features with minimal parameter overhead. The AAM first filters the detailed features through semantic feature ensemble voting and then combines them with semantic features to enhance the semantic information.
Furthermore, we proposed VIM to upsample the downsampled feature to the resolution of the input image. It significantly reduces model latency by only interpolating classes present in the final prediction, which is the most significant contributor to overall model latency. Extensive experiments show that PP-MobileSeg achieves a superior tradeoff between accuracy, model size, and latency compared to other methods. On the ADE20K dataset, PP-MobileSeg achieves 1.57\% higher accuracy in mIoU than SeaFormer-Base with 32.9\% fewer parameters and 42.3\% faster acceleration on Qualcomm Snapdragon 855. Source codes are available at https://github.com/PaddlePaddle/PaddleSeg/tree/release/2.8.
\end{abstract}

\section{Introduction}
\label{sec:intro}

\begin{figure}[tp]
    \begin{center}
      \includegraphics[width=1.0\linewidth]{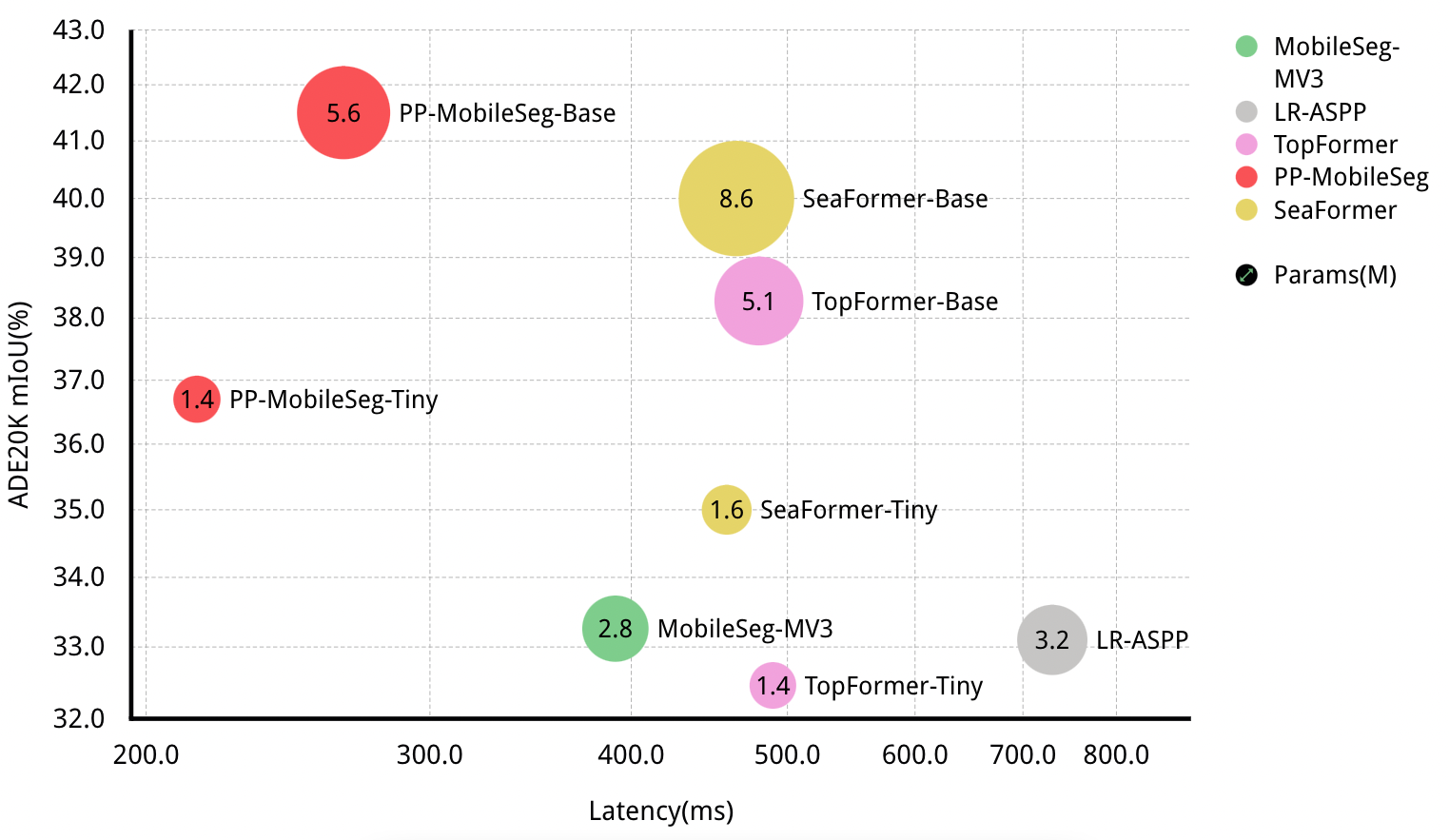}
    \end{center}
    \caption{We present the accuracy-latency-params analysis of our proposed PP-MobileSeg model on the ADE20K validation set. The trade-off analysis is represented as a bubble plot, where the x-axis denotes the latency and the y-axis denotes the mIoU.  Models with the same color are from the same model series. Our model achieves a better accuracy-latency-params trade-off. Note that the latency is tested with the final ArgMax operator using PaddleLite on Qualcomm Snapdragon 855 CPU with a single thread and 512x512 as input shape.}
    \label{fig:tracker}
    \vspace{-0.1in}
\end{figure}

Semantic segmentation is a computationally expensive task compared to other computer vision tasks like image classification~\cite{lu2007survey} or object detection~\cite{zou2023object}, as it involves predicting the class of every pixel. While there have been significant advancements in semantic segmentation on GPU devices, few studies have addressed the challenges of mobile semantic segmentation~\cite{zhang2022topformer, wan2023seaformer, howard2019searching}. This lack of research impedes the practical application of semantic segmentation to mobile applications.

Recently, the surge of vision transformers(ViTs)~\cite{han2022survey} proved the promising performance of transformer-based neural networks on semantic segmentation~\cite{cheng2021per, xie2021segformer, strudel2021segmenter, zheng2021rethinking}. 
Various works have proposed transformer-CNN hybrid architectures for lightweight neural network design, such as MobileViT~\cite{mehta2021mobilevit}, MobileFormer~\cite{chen2022mobile}, and EdgeNext~\cite{maaz2023edgenext}. This hybrid architecture combines global and local information in neural networks at the lowest possible cost. However, the computation complexity of Multi Head Self Attention(MHSA) makes these networks hard to be deployed on mobile devices. Even though several efforts have been made to decrease the time complexity, including shifted window attention~\cite{liu2021swin},  efficient attention~\cite{shen2021efficient},  external attention~\cite{wang2022rtformer}, axial attention~\cite{wang2020axial}, SEA attention~\cite{wan2023seaformer} and etc. But many of these techniques require complex index operations that ARM CPUs cannot support~\cite{wan2023seaformer}. Besides latency and accuracy, memory storage is also a crucial element for mobile applications, since memory storage is limited on mobile devices. Therefore the fundamental question arises: \textit{can we design a hybrid network for mobile devices with a superior trade-off between parameter, latency, and accuracy?}

In this work, we address the above question by exploring the mobile segmentation architecture under the constraints of model size and speed for a performance leap forward. Under extensive search, we manage to propose three novel designed modules: the four-stage backbone StrideFormer, the feature fusion block AAM, and the upsample module VIM as shown in Fig.~\ref{fig:network}. By combining these modules, we propose a family of SOTA mobile semantic segmentation networks called PP-MobileSeg, which is well-suited for mobile devices with great parameters, latency, and accuracy balance. Our improved network design allows PP-MobileSeg-Base to improve the inference speed by 40\% and 34.9\% less model size than SeaFormer while maintaining a competitive 1.37 higher mIoU(Tab.~\ref{overallres}). Compared with MobileSeg-MV3, PP-MobileSeg-Tiny achieves 3.13 higher mIoU while being 45\% faster and 49.5\% smaller(Tab.~\ref{overallres}). We also evaluate the performance of PP-MobileSeg on the Cityscapes dataset\cite{cordts2016cityscapes} (Tab.\ref{cityscapesres}), which shows its superiority in model performance on high-resolution inputs. Although PP-MobileSeg-Base has slightly longer latency, it maintains model size superiority while being 1.96 higher in mIoU than SeaFormer\cite{wan2023seaformer} on the cityscapes dataset~\cite{cordts2016cityscapes}.

In summary, our contributions are as follows:
\begin{itemize}
    \item 
    We introduce the StrideFormer, a four-stage backbone with MobileNetV3 blocks that efficiently extracts features of different receptive fields while minimizing parameter overhead. We also apply strided SEA attention~\cite{li2022efficientformer, wan2023seaformer} to the output of the last two stages to improve global feature representation under computation constraints.
    
    \item We propose the Aggregate Attention Module (AAM), which fuses features from the backbone through ensemble voting of enhanced semantic features and further enhances the fused feature with the semantic feature of the largest receptive field.

    \item To reduce the significant latency caused by the final interpolation and ArgMax operation, we design the Valid Interpolate Module (VIM) that only upsamples classes present in the final prediction during inference time. Replacing the final interpolation and ArgMax operation with VIM significantly reduces model latency.

    \item We combine the above modules to create a family of SOTA mobile segmentation models called PP-MobileSeg. Our extensive experiments show that PP-MobileSeg achieves an excellent balance between latency, model size, and accuracy across ADE20K and Cityscapes datasets.
    
\end{itemize}

\section{Related Work}
Under the speed and model size constraints, mobile semantic segmentation is the task that aims to adapt semantic segmentation networks with efficient network designs.

\subsection{semantic segmentation}
To achieve high performance in semantic segmentation, several key elements are essential, including a large receptive field to capture context~\cite{zhou2017scene, chen2018encoder}, a large resolution of features for accurate segmentation~\cite{wang2020deep,yuan2021hrformer}, fusion of detail and semantic features for precise predictions~\cite{ronneberger2015u, badrinarayanan2017segnet}, and attention mechanisms for improving feature representation~\cite{vaswani2017attention, liu2021polarized, xie2021segformer}. State-of-the-art models often combine several or even all of these elements to achieve superior performance.  The primary requirement for the semantic segmentation task is that the network must be able to capture a holistic view of the scene while simultaneously preserving the image's details and semantics. Thus, it is essential to design network architectures that can efficiently and effectively integrate these elements.

\subsection{Efficient Network Designs}
There are two types of efficient network architectures in the field of deep learning. The first type focuses on adding new elements to the network without introducing unwanted latency during inference. The representative one is structural reparameterization~\cite{ding2022re, ding2021repvgg}, which approximates the multi-branch neural network block with a single branch at inference time. The second type aims to downscale the network at the expense of the model performance reduction. Designs belonging to this category include group convolution~\cite{han2020ghostnet}, channel shuffle~\cite{zhang2018shufflenet}, and efficient attention mechanisms~\cite{shen2021efficient, wang2022rtformer, wang2020axial}.

\subsection{Mobile semantic segmentation}
Due to the large computational complexity of semantic segmentation, there has been limited research of segmentation on mobile devices, with only a few works focusing on this area~\cite{wan2023seaformer, zhang2022topformer, howard2019searching}. Among them, TopFormer enhances the token pyramid with a self-attention block and fuses it with the local feature using their proposed injection module. Further, SeaFormer boosts the model performance with an efficient SEA attention module. Both of them significantly outperform MobileSeg and LRASPP, which currently represent the state-of-the-art in mobile semantic segmentation.

\begin{figure*}[th]
  \centering
      \includegraphics[width=1.0\linewidth]{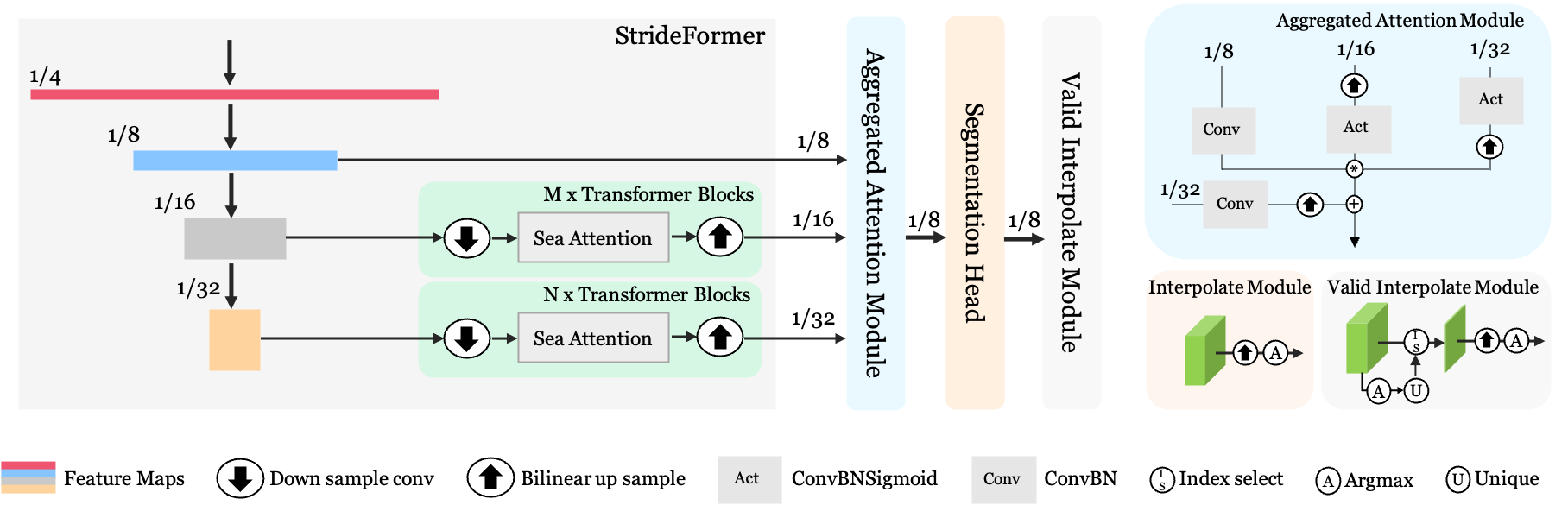}
  \caption{The architecture of PP-MobileSeg network. The structure of AAM is on the top right of the figure. The difference between the normal interpolation module and VIM is displayed at the bottom right of the figure. By selecting only the classes that exist in the final prediction, VIM significantly reduces latency by upsampling a few channels.}
  \label{fig:network}
\end{figure*}

\section{Architecture}
This section presents a comprehensive exploration of mobile segmentation networks designed under speed and size constraints, aiming at achieving better segmentation accuracy. Through our research, we have identified three key modules that lead to faster inference speed or smaller model size with slight performance improvements. The full architecture of PP-MobileSeg is shown in Fig.~\ref{fig:network}, which comprises four main parts: StrideFormer, Aggregate Attention Module (AAM), segmentation head, and Valid Interpolate Module (VIM). The StrideFormer takes input images and generates a feature pyramid with strided attention applied to the last two stages to incorporate global semantics. The AAM is responsible for fusing local and semantic features, which are then passed through the segmentation head to produce the segmentation mask. Finally, the upsample module VIM is used to further enhance the segmentation mask, reducing latency by only upsampling the few channels corresponding to the classes that exist in the final prediction. The following sections provide a detailed description of each of these modules.

\subsection{StrideFormer}
In the StrideFormer module, we utilize a stack of MobileNetV3\cite{howard2019searching} blocks to extract features of different receptive fields. More detailed information about the variants of this architecture can be found in subsection~\ref{variants}. Given an image of \(I \in R^{3\times H \times W}\), where \(3, H, W\) represent channels, height, and width of the image. StrideFormer produces features \(\{F_{\times8}, F_{\times16}, F_{\times32}\}\), representing features that are downsampled 8, 16, and 32 times compared to the resolution of the input image. One key design choice is the number of stages in the backbone, where each stage is a stack of mobilenetv3 blocks that produce one of the feature sets, \(F_{\times downsample-rate}\).  Inspired by efficientFormer\cite{li2022efficientformer}, we discover that the four-stage model has minimal parameter overhead while still maintaining excellent performance compared to the five-stage model as shown in Tab.~\ref{ablation1}. Therefore we design StrideFormer with the four-stage paradigm. With \(\{F_{\times8}, F_{\times16}, F_{\times32}\}\) generated from the four-stage backbone, we add the \(M/N\) SEA attention blocks on the features from the last two stages following \cite{wan2023seaformer}. Due to the time complexity of the self-attention module with large resolution input, we add the stride convolution prior to the SEA attention module and upsample the feature afterward. In this way, we reduce computation complexity by 1/4 of the original implementation when we empower the features with global information.

\subsection{Aggregated Attention Module}
With the generated \(\{F_{\times8}, F_{\times16}, F_{\times32}\}\) from backbone, we designed a aggregated attention module(AAM) to fuse features. The structure of AAM is on the top-right of Fig.~\ref{fig:network}. Among the generated features, \(\{F_{\times16}, F_{\times32}\}\) have a larger receptive field and contain rich semantic information. Therefore we use them as the information filter through ensemble voting to find out the important  information in detail feature \(F_{\times8}\). In the filtration process, \(F_{\times16}\) and \(F_{\times32}\) are upsampled to the same resolution as \(F_{\times8}\). And sigmoid operator is applied to them to obtain weight coefficients.  Afterward, \(F_{\times16}\) and \(F_{\times32}\) is multiplied and the multiplication result is used to filter \(F_{\times8}\). We can formulate the above procedure as Eq.~\ref{eq:aam}

Additionally, we observed that the features with rich semantics complement the previously filtered detail feature and are crucial in improving model performance.  Therefore, it should be kept to the most extent. So we add \(F_{\times32}\), the feature of the largest receptive field and enhanced with the global view, to the filtered detail feature.

\begin{equation}
\begin{split}
  F_{fused} &= Act(F_{\times32}) \times Act(F_{\times16}) \times Conv(F_{\times8}) \\
       &+ Conv(F_{\times32})
\end{split}
  \label{eq:aam}
\end{equation}

After feature fusion, the fused feature captures both rich spatial and semantic information, which is fundamental for segmentation performance.
On top of that, we add a simple segmentation head following TopFormer~\cite{zhang2022topformer}. The segmentation head consists of a \(1\times1\) layer, which helps to exchange information along the channel dimension. Then a dropout layer and convolutional layer are applied to produce the downsampled segmentation map.

\begin{table*}[tp]
\begin{center}
\resizebox{0.9\textwidth}{!}{
\begin{tabular}{ llccc  }
\hline
Method & Backbone & mIoU(\%) & Latency(ms) & Parameters (M)  \\  \hline
LR-ASPP~\cite{howard2019searching} & MobileNetV3-large-x1 & 33.10 & 730.9 & 3.20 \\
MobileSeg~\cite{liu2021paddleseg} & MobileNetV3-large-x1 & 33.26 & 391.5 & 2.85 \\
\hline
TopFormer-Tiny~\cite{zhang2022topformer} & TopTransFormer-Tiny & 32.46 & 490.3 & \textbf{1.41} \\
SeaFormer-Tiny~\cite{wan2023seaformer} & SeaFormer-Tiny & 35.00 & 459.0 & 1.61 \\
\textbf{PP-MobileSeg-Tiny} & StrideFormer-Tiny & \textbf{36.39} & \textbf{215.3} & 1.44\\ \hline
TopFormer-Base~\cite{zhang2022topformer} & TopTransformer-Base & 37.80 & 480.6 & \textbf{5.13} \\
SeaFormer-Base~\cite{wan2023seaformer} & Seaformer-Base & 40.20 & 465.4 & 8.64 \\
\textbf{PP-MobileSeg-Base} & StrideFormer-Tiny & \textbf{41.57} & \textbf{265.5} & 5.71 \\ \hline
\end{tabular}
}
\end{center}
\caption{Results on ADE20K validation set. Latency is measured with PaddleLite with the final ArgMax operator on Qualcomm Snapdragon 855 CPU) and 512x512 as the input shape. All result is evaluated with a single thread. The mIoU is reported with single-scale inference.}
\label{overallres}
\end{table*}

\subsection{Valid Interpolate Module}
Under the latency constraints, we did a latency profile and find out that the final interpolation and ArgMax operation take up more than 50\% of the overall latency. Therefore, we designed the Valid Interpolate Module(VIM) to replace the interpolation and ArgMax operation and greatly reduce the model latency. The latency profile of the SeaFormer-Base and PP-MobileSeg-Base is shown in Fig~\ref{fig:latency}. Detailed statistics after adding VIM can be seen in Table.~\ref{ablation1}.

The VIM is based on the observation that the number of classes that appear in the prediction of a well-trained model is often much less than the overall number of classes in the dataset, especially for datasets with a large number of classes. This is a common case for datasets with a large number of classes. Therefore, it is not necessary to consider all of the classes in the interpolation and ArgMax process. The structure of VIM is on the bottom-right of Fig.~\ref{fig:network}. As the structure shows, the VIM consists of three main steps. First, the ArgMax and Unique operations are applied to the downsampled segmentation map to find out the necessary channels. Then, the index select operation selects only those valid channels, and interpolation is applied to the slimmed feature. Finally, the selected channels are upsampled to the original resolution to produce the final segmentation map. With VIM in replace of interpolation and ArgMax operation, we retrieved the final segmentation map at much less latency costs.

The use of VIM greatly reduces the channels involved in the interpolation and ArgMax operation, leading to a significant decrease in the model latency. However, VIM is only applicable when the number of classes is large enough to have channel redundancy in the model. Therefore, a class threshold of 30 is set, and VIM will not take effect when the number of classes is below this threshold.

\subsection{Architecture Variants}
\label{variants}
We provide two variants of PP-MobileSeg to adapt our model to different complexity requirements, i.e., PP-MobileSeg-Base and PP-MobileSeg-Tiny. The size and latency of these two variants with the input of shape 512x512 are shown in Tab.~\ref{overallres}.  The base and tiny model have the same number of MobileNetV3 layers, whereas the base model is wider than the tiny model. And the base model generates features with more channels to enrich the feature representation. Besides, there are several differences in the attention block as well. PP-MobileSeg-Base model has 8 heads in the SEA attention module, \(M/N=3/3\) attention blocks. The PP-MobileSeg-Tiny model has 4 heads in the SEA attention module and the number of blocks is \(M/N=2/2\). The feature channels of the last two stages are {128, 192} for PP-MobileSeg-Base and {64, 128} for PP-MobileSeg-Tiny respectively. The setting of the feature fusion module is the same for the base and tiny model and the embed channel dim of AAM is set to 256. For more details about the network architecture, please refer to the source code.

\section{Experiments}
In this section, we first present the dataset used for our model training and evaluation and provide implementation details on our training and inference implementation. Secondly, We compared the proposed method with the previous state-of-the-art on this task in terms of accuracy, inference speed, and model size. Finally, we perform an ablation study to demonstrate the effectiveness of our proposed method.

\subsection{Experiments Setup}
\subsubsection{Datasets}
We perform our experiments on ADE20K\cite{zhou2017scene} and cityscapes\cite{cordts2016cityscapes} datasets, and the mean of class-wise intersection over union(mIoU) is used as the evaluation metric. \textbf{ADE20K} is a parsing dataset that contains 25K images in total and 150 fine-grained semantic concepts. All images are split into 20K/2K/3K for training, validation, and testing.
\textbf{Cityscapes} is a large-scale dataset for semantic segmentation. It consists of 5,000 fine annotated images, 2975 for training, 500 for validation, and 1525 for test-dev. The resolution of the images is 2048 × 1024, which poses a great challenge for models used in mobile devices.
\subsubsection{Implementation Details}
Our implementation is built upon PaddleSeg~\cite{liu2021paddleseg} and PaddlePaddle~\cite{ma2019paddlepaddle}. 
\begin{table*}[ht]
\begin{center}
\resizebox{0.8\textwidth}{!}{
\begin{tabular}{ llccc  }
\hline
Method & Backbone & mIoU(\%) & Latency(ms) & Parameters (M)  \\  
\hline
SeaFormer-Small~\cite{wan2023seaformer} & SeaFormer-Small & 70.70 & 204.9 & 1.61 \\
\textbf{PP-MobileSeg-Tiny} & StrideFormer-Tiny & \textbf{70.82} & \textbf{158.3} & \textbf{1.44} \\ \hline
SeaFormer-Base~\cite{wan2023seaformer} & Seaformer-Base & 72.20 & \textbf{297.3} & 8.64 \\
\textbf{PP-MobileSeg-Base} & StrideFormer-Base & \textbf{74.14} & 323.7 & \textbf{5.71} \\ \hline
\end{tabular}}
\end{center}
\caption{Results on Cityscapes validation set. We measure latency using PaddleLite with the final ArgMax operator on an ARM-based device
with a single Qualcomm Snapdragon 855 CPU, with an input shape of 1024x512. As the number of classes in the Cityscapes dataset is small, we did not use the Valid Interpolate Module (VIM). All results are evaluated using a single thread, and we report the mean Intersection over Union (mIoU) using single-scale inference.}
\label{cityscapesres}
\end{table*}

\begin{figure}
  \centering
  \begin{subfigure}{0.47\linewidth}
   \includegraphics[width=1.0\linewidth]{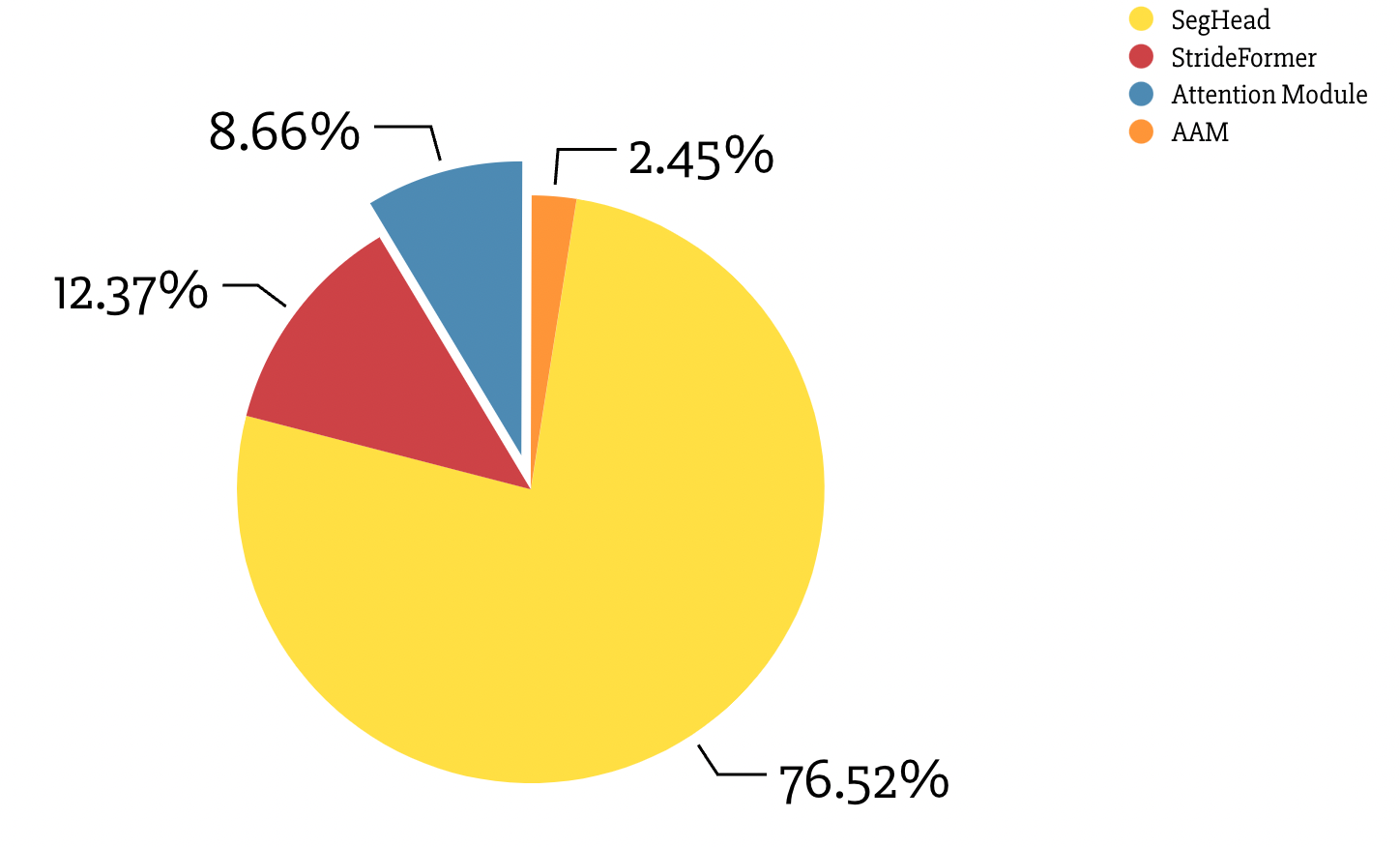}
    \caption{The latency profile of SeaFormer~\cite{wan2023seaformer}.}
    \label{fig:short-a}
  \end{subfigure}
  \hfill
  \begin{subfigure}{0.47\linewidth}
    \includegraphics[width=1.0\linewidth]{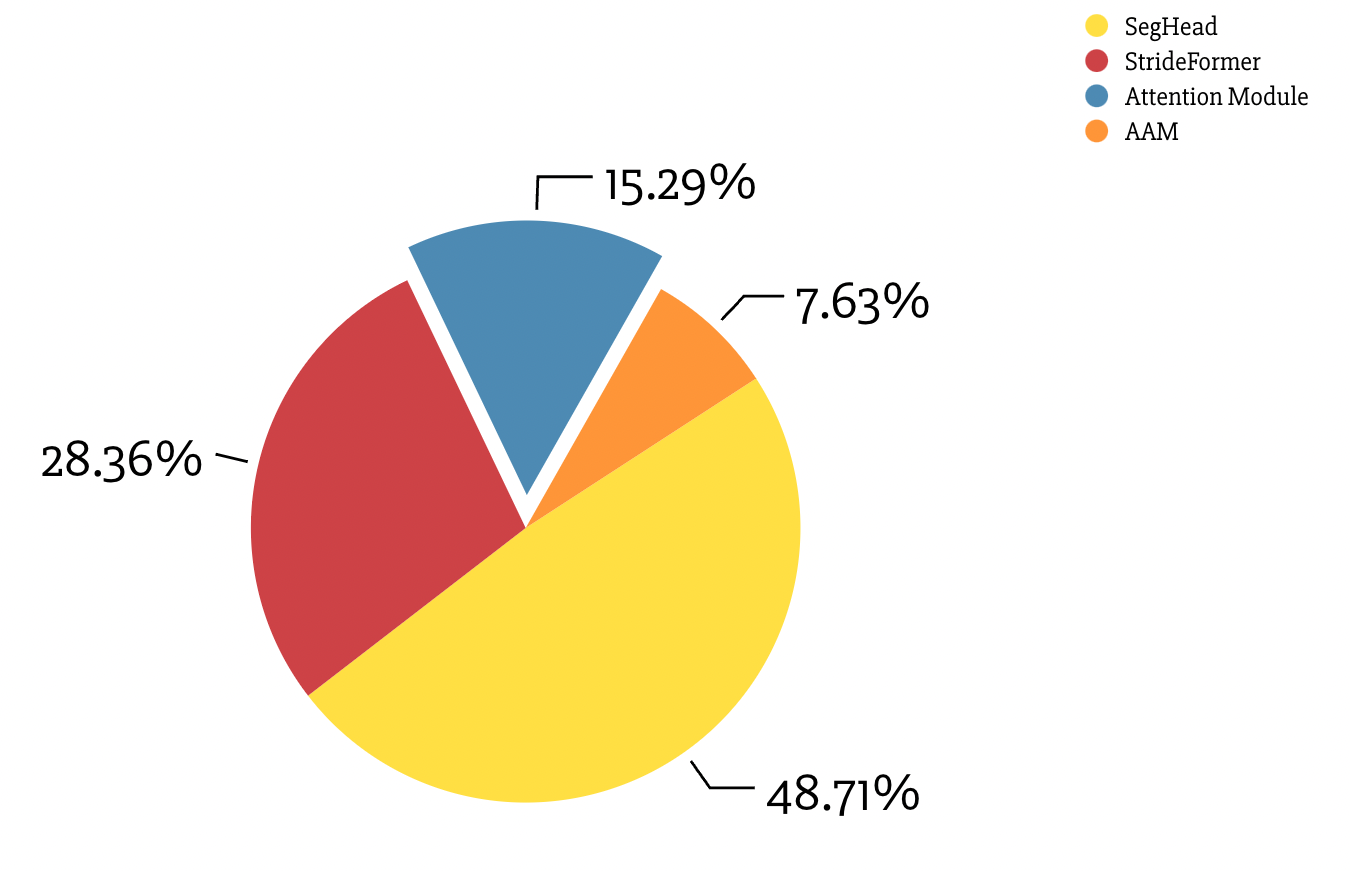}
    \caption{The latency profile of PP-MobileSeg.}
    \label{fig:short-b}
  \end{subfigure}
  \caption{Latency profile compare between SeaFormer and PP-MobileSeg.}
  \label{fig:latency}
\end{figure}

\begin{table}[ht]
    \begin{center}
        \resizebox{0.48\textwidth}{!}{
        \begin{tabular}{ ccc|ccc}
        \hline
        VIM & StrideFormer & AAM & mIoU & Latency(ms) & Params(M)\\  \hline
         & & & 40.20 & 465.6 & 8.27 \\
        $\surd$ & & & 40.20 & \textbf{234.6} & 8.17 \\
        $\surd$ & $\surd$ & & 40.98 & 235.1 & \textbf{5.54} \\
        $\surd$ & $\surd$ &$\surd$ & \textbf{41.57} & 265.5 & 5.71 \\\hline
        \end{tabular}}
    \end{center}
\caption{Ablation Study of three proposed modules of PP-MobileSeg-Base on ADE20K dataset.}
\label{ablation1}
\end{table}

\textbf{Training Settings}
Our backbone is pre-trained on ImageNet1K~\cite{deng2009imagenet} to retrieve common knowledge about images. We set the batch size as 32 and train the model for 80K iterations. During training, we cross entropy loss and Lovasz loss with the loss ratio of 4:1~\cite{berman2018lovasz} to optimize the model. We use the exponential moving average method to average model parameters from different training iterations and the moving average coefficient is 0.999~\cite{tarvainen2017mean}. The learning rate is at 0.006 and uses the ADAMW~\cite{loshchilov2017decoupled} optimizer with the weight decay set at 0.01. the learning rate schedule is set as the combination of the warmup schedule and the poly schedule with factor 1.0. The learning rate goes up from 1e-6 for 1500 iters and then decreases linearly. For ADE20K, we follow the data augmentation strategy of TopFormer and SeaFormer\cite{zhang2022topformer, wan2023seaformer}, including the random scale ranges in [0.5, 2.0], image crop to the given size, random horizontal flip, and random distortion. For Cityscapes, the data augmentation is the same except that we crop the image to 1024x512 rather than 512x512 in ADE20K datasets, and the random scale ranges in [0.125, 1.5]. Our model is trained with two Tesla V100 GPUs. We report the single scale results on the validation set to compare with other methods.

\textbf{Inference Settings}
During inference, we set the input shape as \(512\times512\)  for ADE20K datasets and \(512\times1024\) for cityscapes.  To test the model latency, the full-precision PP-MobileSeg models are exported to the static model, and the latency is then measured on the Qualcomm 855 with PaddleLite on a single thread. During inference, we use VIM in place of the interpolation and ArgMax operation. It is worth noting that the  image preprocesses, including resizing and normalizing, is accomplished before the inference process, so the inference time only includes model infer time. Especially, the latency of VIM is correlated with the number of classes predicted in the image. Therefore, we use an image from the ADE20K validation set, which has the average number of categories, to evaluate the latency for a reasonable comparison.

\subsection{Comparison with State-of-the-arts}
\textbf{ADE20K Results}
Table.~\ref{overallres} presents the comparison of PP-MobileSeg with previous mobile semantic segmentation models, including both lightweight vision transformers and efficient CNNs, and report the results on parameters, latency, and mIoU. As the results show, PP-MobileSeg outperforms these SOTA models not only on latency but also on model size while maintaining a competitive edge in accuracy. Compare with MobileSeg and LRASPP, both of them use MobileNetV3 as their backbone, PP-MobileSeg-Tiny is more than 3.0 higher than them in mIoU, while being 49.47\% smaller and 45\% faster than MobileSeg. And PP-MobileSeg-Tiny is 55\% smaller and 70.5\% faster than LRASPP. In comparison with the SOTA vision transformer-based models TopFormer and SeaFormer, which use convolution-based global self-attention as their semantics extractor,  PP-MobileSeg achieves higher segmentation accuracy with lower latency and smaller model size.  PP-MobileSeg-Base is about the same size or 34.9\% smaller and 42.9\% to 44.7\% faster than its counterparts, while maintaining a competitive edge in accuracy and is 1.37\% to 3.77\% higher in mIoU. These results demonstrate the effectiveness of PP-MobileSeg in improving feature representation.

\textbf{Cityscapes Results}
It can be seen from Table.~\ref{cityscapesres} that PP-MobileSeg-Tiny achieves better performance in all aspects of accuracy, latency, and parameters than SeaFormer-small. Furthermore, PP-MobileSeg-Base achieves significantly better accuracy with comparable latency and smaller model sizes. These results demonstrate that PP-MobileSeg maintains its excellent balance among accuracy, model size, and speed even under high-resolution inputs.

\subsection{Ablation Study}
We conduct an ablation study to discuss the influence of our proposed modules and dissect and analyze these modules. In Table~\ref{ablation1}, we show the effectiveness of three proposed modules by adding them to the baseline one by one. 

\textbf{VIM}: As we mentioned before, VIM serves as the replacement for interpolation and ArgMax operations to accelerate the inference speed. As we can see from the profile comparison (Fig.~\ref{fig:latency}), the overall latency of Segmentation greatly decreased from 76.32\% to 48.71\% with the application of VIM. And the experimental results from Table~\ref{ablation1} show the model latency is decreased by 49.5\% after adding VIM. These experiments prove that VIM's acceleration capabilities on datasets with a large number of classes are exceptional.

\begin{table}[ht]
    \begin{center}
        \resizebox{0.48\textwidth}{!}{
        \begin{tabular}{ cc|cc}
        \hline
        ensemble vote & final semantics & mIoU  & Params(M)\\  \hline
        $\surd$ & &  41.12  & 5.71 \\
         & $\surd$ & 41.53  & \textbf{5.63} \\
        $\surd$ & $\surd$ & \textbf{41.57} & 5.71 \\\hline
        \end{tabular}}
    \end{center}
    \caption{Ablation Study of the AAM module. Ensemble vote represents the multiplication of \(F_{\times 32}\) and \(F_{\times 16}\) to filter \(F_{\times 8}\), without it, \(F_{\times 16}\) and \(F_{\times 32}\) will filter  \(F_{\times 8}\) the one by one as SeaFormer~\cite{wan2023seaformer}. Final semantics represent whether we add \(F_{\times 32}\) to the fused feature.}
    
    \label{ablation2}
\end{table}

\textbf{StrideFormer}: The usage of the four-stage network in StrideFormer resulted in a notable reduction of 32.19\% in parameter overhead. The experimental results also show an increase in accuracy by 0.78\%, which we attribute to the enhanced backbone.

\textbf{AAM}: AAM raises the accuracy by 0.59\% while slightly increasing the latency and model size. To gain insight into the design of the AAM, we split the fusion module into two branches: the ensemble vote and the final semantics as shown in Table~\ref{ablation2}. And the reported results reveal the significance of both branches and especially the importance of final semantics. Without it, the accuracy can drop by 0.45\%.

\section{Conclusion}
In this paper, we investigate the design options for hybrid vision backbones and addressed the latency bottleneck in mobile semantic segmentation networks. After thorough exploration, we identified the mobile-friendly design choices and propose a new family of mobile semantic segmentation networks called PP-MobileSeg with the combination of transformer blocks and CNN. With the carefully designed backbone, fusion module and interpolate module, PP-MobileSeg achieves a SOTA balance between model size, speed, and accuracy on ARM-based devices. 

{\small
\bibliographystyle{ieee_fullname}
\bibliography{egbib}
}

\end{document}